\documentclass[sigconf, nonacm]{acmart}

\newcommand\vldbdoi{XX.XX/XXX.XX}
\newcommand\vldbpages{XXX-XXX}
\newcommand\vldbvolume{14}
\newcommand\vldbissue{1}
\newcommand\vldbyear{2020}
\newcommand\vldbauthors{\authors}
\newcommand\vldbtitle{\shorttitle} 
\newcommand\vldbavailabilityurl{URL_TO_YOUR_ARTIFACTS}
\newcommand\vldbpagestyle{plain}

\usepackage{amsmath,amsfonts}
\usepackage{algorithmic}
\usepackage{algorithm}
\usepackage{array}
\usepackage[caption=false]{subfig}
\usepackage{textcomp}
\usepackage{bm}
\usepackage{stfloats}
\usepackage{url}
\usepackage{verbatim}
\usepackage{graphicx}
\usepackage{subfiles}
\usepackage{multirow}

\begin{document}

\title{PAST: A Primary-Auxiliary Spatio-Temporal Network for Traffic Time Series Imputation}

\author{Hanwen Hu}
\orcid{0000-0002-1825-0097}
\affiliation{
  \institution{Shanghai Jiao Tong University}
  \streetaddress{Dongchuan Road 800}
  \city{Shanghai}
  \postcode{200240}
}
\email{hanwen\_hu@sjtu.edu.cn}

\author{Zimo Wen}
\orcid{0000-0002-1825-0097}
\affiliation{
  \institution{Shanghai Jiao Tong University}
  \streetaddress{Dongchuan Road 800}
  \city{Shanghai}
  \postcode{200240}
}
\email{2581235653@sjtu.edu.cn}

\author{Shiyou Qian}
\orcid{0000-0001-5109-3700}
\affiliation{
  \institution{Shanghai Jiao Tong University}
  \streetaddress{Dongchuan Road 800}
  \city{Shanghai}
  \postcode{200240}
}
\email{qshiyou@sjtu.edu.cn}

\author{Jian Cao}
\orcid{0000-0001-5109-3700}
\affiliation{
  \institution{Shanghai Jiao Tong University}
  \streetaddress{Dongchuan Road 800}
  \city{Shanghai}
  \postcode{200240}
}
\email{cao-jian@sjtu.edu.cn}

\begin{abstract}
Traffic time series imputation is crucial for the safety and reliability of intelligent transportation systems, while diverse types of missing data, including random, fiber, and block missing make the imputation task challenging. Existing models often focus on disentangling and separately modeling spatial and temporal patterns based on relationships between data points. However, these approaches struggle to adapt to the random missing positions, and fail to learn long-term and large-scale dependencies, which are essential in extensive missing conditions. 
In this paper, patterns are categorized into two types to handle various missing data conditions: primary patterns, which originate from internal relationships between data points, and auxiliary patterns, influenced by external factors like timestamps and node attributes. Accordingly, we propose the \textbf{P}rimary-\textbf{A}uxiliary \textbf{S}patio-\textbf{T}emporal network (PAST). It comprises a graph-integrated module (GIM) and a cross-gated module (CGM). GIM captures primary patterns via dynamic graphs with interval-aware dropout and multi-order convolutions, and CGM extracts auxiliary patterns through bidirectional gating on embedded external features. The two modules interact via shared hidden vectors and are trained under an ensemble self-supervised framework. Experiments on three datasets under 27 missing data conditions demonstrate that the imputation accuracy of PAST outperforms seven state-of-the-art baselines by up to 26.2\% in RMSE and 31.6\% in MAE.
\end{abstract}

\maketitle

\pagestyle{\vldbpagestyle}
\begingroup\small\noindent\raggedright\textbf{PVLDB Reference Format:}\\
\vldbauthors. \vldbtitle. PVLDB, \vldbvolume(\vldbissue): \vldbpages, \vldbyear.\\
\href{https://doi.org/\vldbdoi}{doi:\vldbdoi}
\endgroup
\begingroup
\renewcommand\thefootnote{}\footnote{\noindent
This work is licensed under the Creative Commons BY-NC-ND 4.0 International License. Visit \url{https://creativecommons.org/licenses/by-nc-nd/4.0/} to view a copy of this license. For any use beyond those covered by this license, obtain permission by emailing \href{mailto:info@vldb.org}{info@vldb.org}. Copyright is held by the owner/author(s). Publication rights licensed to the VLDB Endowment. \\
\raggedright Proceedings of the VLDB Endowment, Vol. \vldbvolume, No. \vldbissue\ %
ISSN 2150-8097. \\
\href{https://doi.org/\vldbdoi}{doi:\vldbdoi} \\
}\addtocounter{footnote}{-1}\endgroup

\ifdefempty{\vldbavailabilityurl}{}{
\begingroup\small\noindent\raggedright\textbf{PVLDB Artifact Availability:}\\
The source code and data have been made available at \url{https://github.com/Hanwen-Hu/PAST}.
\endgroup}

\section{Introduction}
Traffic time series are essential for the development of intelligent transportation systems \cite{me}. However, missing data caused by sensor failures \cite{sensor}, transmission errors \cite{transmission}, or environmental disturbances may undermine the reliability of analytical outcomes or the efficacy of real-time decision-making. For example, in real-time traffic management, data loss can render adaptive signal control systems unresponsive to traffic flow fluctuations \cite{signal}. Furthermore, missing data may delay incident detection and alerts, thereby extending emergency response times and elevating the risk of secondary accidents. In traffic planning, prolonged data gaps would distort assessments of road network capacity, leading to flawed decisions on infrastructure expansion \cite{capacity}. Therefore, precise imputation of the gaps in data is critical for traffic systems.

\begin{figure*}
    \centering
    \includegraphics[width=\linewidth]{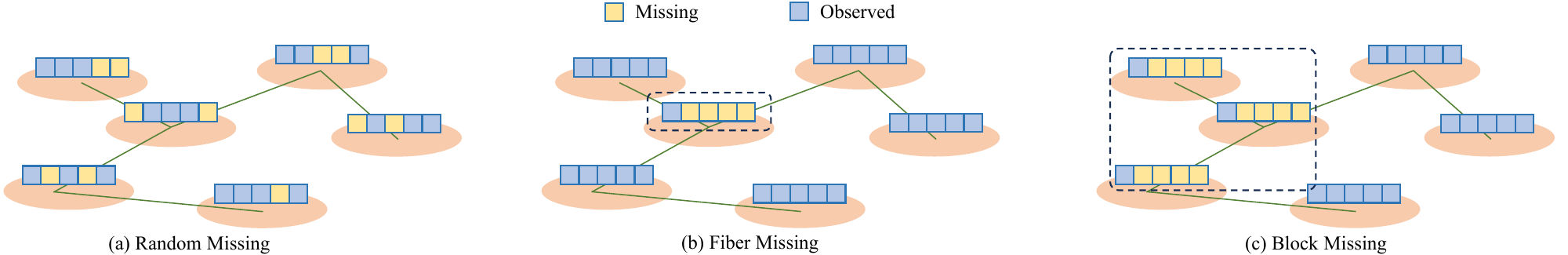}
    \caption{Three types of missing data in traffic scenarios.}
    \Description{}
    \label{fig:miss_intro}
\end{figure*}

Previous studies classified missing data in traffic scenarios into three types \cite{survey}, as shown in Figure \ref{fig:miss_intro}: (1) random missing, resulting from sporadic transmission errors, where missing entries are independently distributed across time and space; (2) fiber missing, caused by prolonged sensor disconnection, leading to extended temporal gaps at specific locations; and (3) block missing, arising from the absence of sensors over a large region \cite{block_miss}, leading to spatially and temporally contiguous gaps in the collected data.

Accurate traffic data imputation faces two challenges. First, heterogeneous spatio-temporal patterns interact distinctly in various missing scenarios. For instance, random missing is dominated by short-term fluctuations, whereas fiber missing relies mainly on long-term periodicity, and block missing requires capturing large-scale spatial dependencies. 
Second, unlike autoregressive forecasting, imputation lacks ground truth for missing values, and the randomness of missing positions leads to overfitting on spurious correlations. Static architectures cannot dynamically learn dependencies based on observed contexts, leading to biased estimations \cite{irnet}.

Researchers have proposed various innovative approaches to tackle these challenges, which can be broadly classified into three main categories.
First, classical models such as K-Nearest Neighbors (KNN) \cite{KNN}, linear interpolation, MICE \cite{mice}, and matrix factorization \cite{TIDER} are computationally efficient, but struggle to learn complex relationships. 
Second, deep learning models including BRITS \cite{BRITS}, GAIN \cite{GAIN}, IR$^2$-Net \cite{irnet}, and GRIN \cite{GRIN}, employ various self-supervised frameworks for time series imputation. However, they are primarily designed for randomly missing and often fail to generalize to fiber or block missing patterns-—which are prevalent in real-world traffic monitoring systems. 
Third, traffic-specific models such as STCPA \cite{STCPA} and SAGCIN \cite{SAGCIN}, account for multiple missing types but rely solely on internal patterns between observed data points, without explicitly aligning pattern utilization with specific missing scenarios. 
Consequently, their performance degrades under extensive or structured missing conditions due to the reliance on pattern learning between local data points, necessitating a more adaptive and comprehensive framework.

To address these limitations, we propose to distinguish latent patterns in traffic data as either primary or auxiliary. 
Primary patterns, arising from relationships among data points within a time series and being considered in most existing models, can perceive local temporal fluctuations and adjacent topological dependencies. For example, the current data point can generally be estimated from the previous one; changes in upstream nodes in a traffic network would directly impact downstream ones. These patterns are consistent with human intuition but hard to extract when modeling long-term periodicity and spatial similarity.
Therefore, auxiliary patterns, driven by external factors such as timestamps or consistent environmental conditions, are introduced to characterize long-term periodicity and spatial similarity effectively. For instance, morning rush hours stem from fixed travel schedules, not the rush hours in the previous day; and spatial similarity reflects the consistent environmental contexts, rather than direct node interactions. Compared to modeling primary patterns with larger amount of data, concentrating on external information is apparently a more efficient approach to learn long-term and large-scale dependencies. 
In summary, in random missing scenarios, primary patterns are dominant, while auxiliary patterns become important in fiber or block missing scenarios due to their extensive spatial and temporal spans. By integrating auxiliary patterns with primary ones, high-accuracy imputation can be achieved across diverse missing conditions.

Driven by these ideas, we design a primary-auxiliary spatio-temporal network (PAST) for general traffic time series imputation. PAST comprises two modules: a graph-integrated module (GIM) and a cross-gated module (CGM).
GIM is a pure graph neural network (GNN). It captures primary patterns by modeling incomplete time series as a dynamic directed graph, where missing data corresponds to the absence of nodes or edges. An interval-aware dropout mechanism is also designed to improve robustness against varying missing positions.
CGM perceives auxiliary spatio-temporal patterns impact by timestamps and node information, using a gated mechanism to facilitate spatial and temporal pattern interactions and generate effective representations.
The two modules are interconnected via shared hidden vectors, and the model is trained with an ensemble learning framework.

Experiments are conducted on three datasets, namely METR-LA \cite{stgcn}, PeMS-Bay \cite{dcrnn}, and LargeST-SD \cite{largest}, with 9 missing conditions for each dataset. The experimental results demonstrate that the imputation accuracy of PAST outperforms seven state-of-the-art baseline models by up to 26.2\% in RMSE.

The main contributions of this paper are as follows:
\begin{itemize}
    \item We propose PAST, a model for traffic time series imputation, which disentangles spatial and temporal patterns into primary and auxiliary categories, utilizing graph-integrated and cross-gated modules for robust pattern extraction.
    \item We design a graph-integrated module (GIM) to capture primary patterns by modeling incomplete time series as dynamic directed graphs, ensuring stable learning despite varying missing positions.
    \item We develop a cross-gated module (CGM) to extract auxiliary patterns, providing robust support for imputation, especially in fiber and block missing scenarios.
    \item We conduct comprehensive experiments on 27 missing data conditions, comparing PAST with seven time series imputation models to demonstrate its accuracy improvement.
\end{itemize}

\section{Related Work}

\subsection{General Time Series Imputation Models}
Many existing studies address general time series imputation tasks, which are applicable to domains such as industry or meteorology. These models can be broadly classified into traditional machine learning methods and deep learning methods. Traditional models, including linear interpolations, KNN \cite{KNN}, MICE \cite{mice}, and matrix factorization \cite{TIDER}, offer interpretability and efficiency in simple imputation scenarios. In contrast, deep learning models leverage diverse self-supervised frameworks during training. For instance, GAIN \cite{GAIN} employs generative adversarial networks (GAN), where the generator imputes time series and the discriminator differentiates between imputed and observed values. Additionally, GRIN \cite{GRIN} and BRITS \cite{BRITS} are RNN-based models that use bidirectional RNNs to forecast observed values sequentially, thereby learning latent temporal relationships for imputation. Other innovative architectures, such as diffusion-based CSDI \cite{CSDI} and transformer-based SAITS \cite{SAITS}, also achieve strong performance.

Despite the advancements, most of these models primarily address random missing scenarios, and fail to disentangle multiple patterns, limiting their applicability in traffic-specific scenarios.

\subsection{Traffic-Specific Imputation Models}
Building upon general models, several studies tailor imputation methods to traffic data, emphasizing spatio-temporal dependencies. For example, STCPA \cite{STCPA} introduces a cyclic training procedure and a self-attention mechanism for point-by-point imputation, enhancing model stability but overlooking diverse latent patterns. In contrast, some studies incorporate multiple patterns in traffic time series. MDGCN \cite{MDGCN} uses recurrent and graph convolutional layers to capture temporal and spatial information separately, augmented by an external memory network for information storage. GT-TDI \cite{GTTDI} adopts a graph transformer architecture to impute missing values by leveraging semantic understanding of road networks. DCGRIN \cite{DCGRIN} focuses on local correlations and dynamic features through an attention module and spatial-correction graph convolutional layers. Although these studies consider disentangling patterns for traffic imputation, they focus on separating spatial and temporal patterns, instead of adopting external information to enhance performance, and rarely address interactions between patterns. 

\subsection{External Information for Time Series}
Some previous studies highlight the importance of external information in time series analysis. For example, transformer-based forecasting models like Informer \cite{informer}, Autoformer \cite{autoformer} and FEDformer \cite{fedformer} integrate timestamps into positional encoding vectors. STID \cite{STID}, an MLP-based forecasting model, appends spatial and temporal identity information to distinguish similar inputs. Notably, TimeXer \cite{TimeXer} incorporates exogenous variables to forecast endogenous ones via embedding layers and canonical Transformer frameworks. These studies significantly improve the accuracy of time series forecasting. However, external information has not been applied to time series imputation tasks. It typically serves as a supplementary to the inputs, instead of being processed independently by a dedicated module.

\begin{table}[t]
\centering
\caption{Comparison between Imputation Models}
\label{tab:compare}
\begin{tabular}{ccc}\hline
Model&Missing Type&Pattern Type\\\hline
KNN\cite{KNN}&Random&-\\
MICE\cite{mice}&Random&-\\
BRITS\cite{BRITS}&Random&Primary\\
GRIN\cite{GRIN}&Random\&Fiber&Primary\\
CSDI\cite{CSDI}&Random&Primary\\
STCPA\cite{STCPA}&Random&Primary\\
MDGCN\cite{MDGCN}&Random\&Fiber\&Block&Primary\\
GT-TDI\cite{GTTDI}&Random\&Fiber&Primary\\
\textbf{PAST}&Random\&Fiber\&Block&Primary\&Auxiliary\\\hline
\end{tabular}
\end{table}

\subsection{Summary}
Table \ref{tab:compare} compares PAST with several mentioned imputation models. In summary, PAST differs from them in two key aspects. First, instead of modeling entangled patterns with a unified framework, or disentangling them solely along spatial and temporal dimensions, PAST addresses patterns from a primary-auxiliary perspective, enabling explicit features to effectively support imputation in structured missing scenarios. 
Second, PAST establishes correspondences between patterns and missing types, employing dedicated modules to learn these patterns, thereby ensuring generality and stability across diverse missing scenarios. These innovations position PAST as a versatile model for traffic time series imputation, addressing gaps in prior research.

\section{Preliminaries}
The goal of traffic time series imputation is to reconstruct missing values given observed values from a traffic network. Let $\bm{X}=\{\bm{x_1}, \bm{x_2}, \cdots, \bm{x_N}\}\in \mathbb{R}^{L\times N}$ denote the traffic data, where $L$ and $N$ represent the sequence length and the number of nodes, respectively.
$\bm{x_i}=\{x_{1i}, x_{2i}, \cdots, x_{Li}\}\in \mathbb{R}^L$ represents the values of the $i^{th}$ node in the network. A mask matrix $\bm{M}\in \{0,1\}^{L\times N}$ indicates the observed or missing entries in $\bm{X}$:  
\begin{equation}
m_{ij} = \left\{
\begin{aligned}
1,\quad&\text{if $x_{ij}$ is observed,}\\
0,\quad&\text{if $x_{ij}$ is missing.}\\
\end{aligned}
\right.
\end{equation}

The topology of the road network is modeled as a weighted graph $G_\mathcal{S}=<V_\mathcal{S},E_\mathcal{S}>$, where $V_\mathcal{S}$ represents the set of nodes and $E_\mathcal{S}$ denotes the set of weighted edges. The adjacency matrix, denoted as $\bm{A}_\mathcal{S}\in \mathbb{R}^{N\times N}$, represents the weighted connections between nodes in $G_\mathcal{S}$. The objective is to estimate the missing values $\hat{x}_{ij}$ where $m_{ij}=0$, leveraging both observed data and the network’s topological structure. The notations defined in this paper are listed in Table \ref{tab:note}.

\begin{table}[t]
    \centering
    \caption{Notations for the PAST model.}
    \label{tab:note}
    \begin{tabular}{c|c}\hline
    Notations&Descriptions\\\hline
    Sequence Length&$L$\\
    Node Number&$N$\\
    Missing Rate&$r$\\
    Incomplete Time Series&$\bm{X}$\\
    Imputed Time Series&$\bm{Y}$\\
    Spatial Topology&$G_\mathcal{S}=<V_\mathcal{S},E_\mathcal{S}>$\\
    Temporal Topology&$G_\mathcal{T}=<V_\mathcal{T},E_\mathcal{T}>$\\\hline
    \end{tabular}
\end{table}

\section{Methodology}
\subsection{Basic Ideas}
\label{sec:idea}

\begin{figure}[t]
    \centering
    \includegraphics[width=\linewidth]{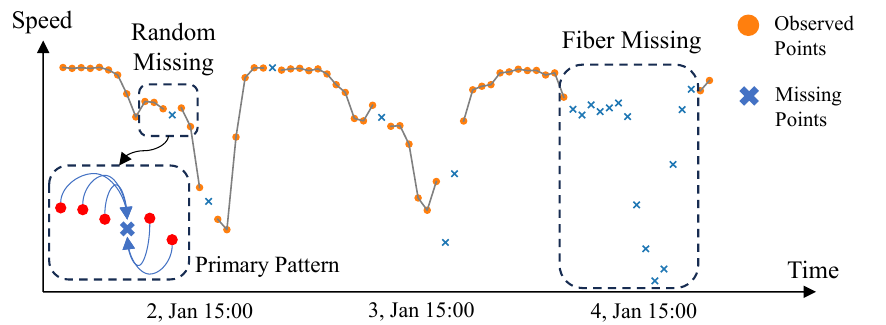}
    \caption{The time series record the speed of a node in the PeMS-Bay dataset \cite{dcrnn}. It provides specific examples of random and fiber missing.}
    \Description{}
    \label{fig:observation}
\end{figure}

Figure \ref{fig:observation} provides a specific example of random and fiber missing. In the case of random missing, absent values can be accurately inferred from adjacent observed values due to consistent local dependencies \cite{irnet}. However, in fiber missing conditions, long-term relationships dominate, making it challenging to capture these dependencies solely from values within the sequence. For example, imputing values on January 4 at 15:00 should consider the historical values from the same period, such as January 3 at 15:00 and January 2 at 15:00. Although modeling longer sequences is a potential solution, it increases computational complexity and reduces efficiency.

Viewed from a causal perspective, the temporal periodicity and spatial characteristics manifested in traffic sequences are driven by external features. For instance, morning rush hours arise from fixed daily travel schedules, while the unique fluctuation patterns of a node stem from its local environmental context. Thus, incorporating external information, such as timestamps and node identities, enables effective capture of long-range and large-scale dependencies without relying on excessively long sequences.

In this regard, we define primary patterns as spatial and temporal dependencies derived purely from internal sequence values, whereas auxiliary patterns refer to those inferred from external features, such as timestamps and node-specific attributes. 
There exists a natural correspondence between missing types and pattern types: random missing mainly relies on extracting primary patterns, whereas fiber and block missing necessitate additional modeling of auxiliary ones to enhance performance. Therefore, a dual-modality framework, comprising primary and auxiliary components, can provide superior generalization across missing types compared to the conventional spatio-temporal modeling paradigms, as it aligns pattern extraction with specific missing scenarios.

\begin{figure}[t]
    \includegraphics[width=\linewidth]{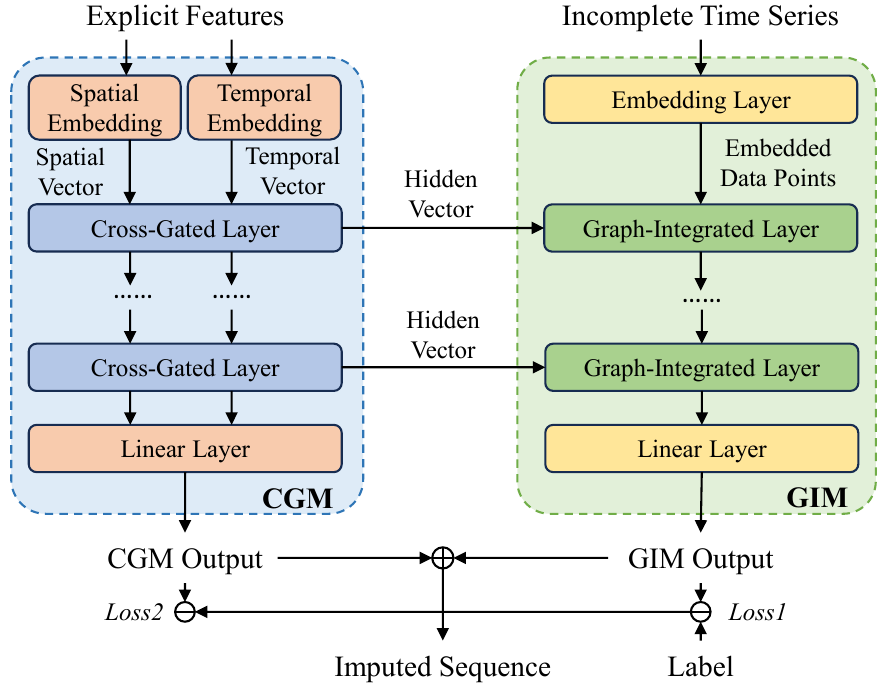}
    \caption{The framework of PAST.}
    \Description{}
    \label{fig:framework}
\end{figure}

Figure \ref{fig:framework} illustrates the framework of the proposed PAST model. It adopts a dual-module architecture: the graph-integrated module (GIM) for primary pattern extraction, and the cross-gated module (CGM) for auxiliary pattern extraction. Each module independently processes its respective patterns, with interactions via hidden vectors that bridge primary and auxiliary representations to emphasize their complementary roles.

The GIM takes as input the incomplete time series $\bm{X}$ and the mask matrix $\bm{M}$. Data points in $\bm{X}$ are first embedded into vectors, then processed through $n$ graph-integrated layers to extract primary patterns. Lastly, a linear layer is employed to yield $\bm{Y}_{GIM}$, representing imputations based on internal data relationships. 
In contrast, the CGM processes external spatial and temporal information, including node identities and timestamps (week, hour, and minute). Inspired by Transformer \cite{transformer}, these features are treated analogously as tokens. Specifically, node identities are embedded to form the spatial vector, while timestamp components are embedded separately and concatenated together to form the temporal vector. These vectors then pass through $n$ cross-gated layers to extract auxiliary patterns, followed by a linear layer to produce $\bm{Y}_{CGM}$. Finally, the imputation result is obtained by fusing the outputs: 
\begin{equation}
    \bm{Y}=\bm{M}\odot\bm{X} + (1-\bm{M})\odot(\bm{Y}_{CGM}+\bm{Y}_{GIM}).
\end{equation}

The interactions between GIM and CGM occur in two aspects. First, each cross-gated layer in CGM forwards a hidden vector encapsulating external information to the corresponding graph-integrated layer in GIM, supplementing primary pattern extraction. Second, inspired by GBDT \cite{gbdt}, GIM is trained with self-supervised mean squared error (MSE) on observed values (Loss 1 in Figure \ref{fig:framework}), while CGM minimizes the MSE between its output and GIM's training residuals.
\begin{equation}
\begin{aligned}
    Loss_1 &= MSE(\bm{M}\odot\bm{Y}_{GIM}, \bm{M}\odot\bm{X})\\
    Loss_2 &= MSE(\bm{M}\odot\bm{Y}_{CGM}, \bm{M}\odot(\bm{Y}_{GIM}-\bm{X}))\\
\end{aligned}
\end{equation}
This residual fitting positions CGM as an auxiliary enhancer, capturing patterns overlooked by GIM and promoting robustness across different missing types.

\subsection{Graph-Integrated Module}
\begin{figure}[t]
    \centering
    \includegraphics[width=\linewidth]{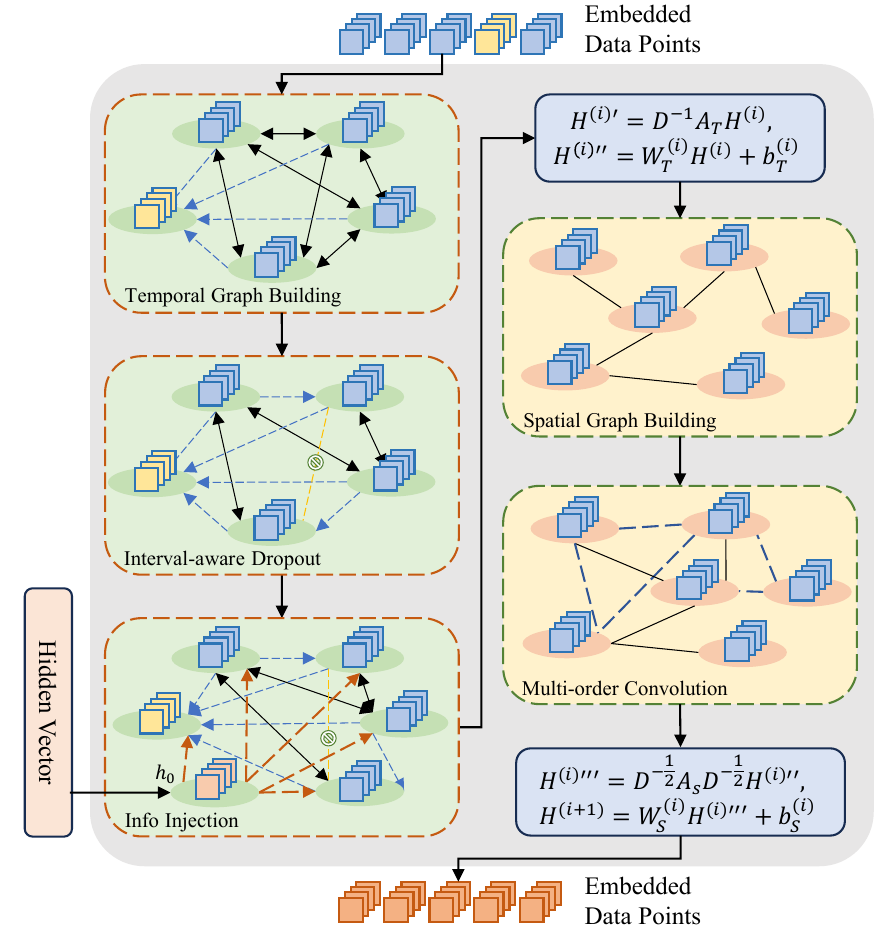}
    \caption{Graph-integrated layer for primary patterns}
    \Description{}
    \label{fig:graph_integrated_unit}
\end{figure}

The Graph Integrated Module (GIM) is designed to capture the primary spatio-temporal patterns, specifically dependencies between data points across time steps and nodes, as defined in the introduction. 
Unlike traditional models that treat missing values as zeros or ignore them, GIM innovatively constructs dynamic graphs to propagate information unidirectionally, ensuring robust pattern extraction integrated with external influences. The following will introduce the core component of GIM: the graph-integrated layer, as shown in Figure \ref{fig:graph_integrated_unit}, which consists of two stacked sub-layers: a temporal layer and a spatial layer.

\subsubsection{Temporal Layer}
The temporal layer models the relationships among data points across different timestamps, containing four key steps: directed graph building, interval-aware dropout, external info injection, and feed forward process.

\paragraph{Temporal Graph Building.}
The layer constructs incomplete time series into a directed graph $G_{\mathcal{T}} = <V_{\mathcal{T}}, E_{\mathcal{T}}>$. The vertex set $V_{\mathcal{T}}$ comprises embedding vectors of $L$ data points in the time series ($L$ is the sequence length). The edge set $E_{\mathcal{T}}$ is defined as follows:
\begin{itemize}
    \item Bidirectional edges between observed values form a fully connected subgraph,  reflecting mutual influences across timestamps.
    \item Unidirectional edges from observed to missing values enable information propagation for imputation.
    \item There is no edge between missing values, as their relationships are inferred from observed data.
\end{itemize}
Most existing studies utilize recurrent neural networks (RNNs), or convolutional neural networks (CNNs) to learn temporal dependencies in time series, regarding them as sequences or figures. However, these two architectures cannot address the randomness of missing positions, or perceive dependencies from a whole scale of sequence. Therefore, the GNN architecture is chosen to learn temporal dependencies, which can dynamically adapt to the change of missing patterns by modifying the status of nodes and edges, enabling robust information propagation from observed to missing entries.

\paragraph{Interval-aware Dropout.} To mitigate overfitting caused by the variability of missing positions and enhance stability, an interval-aware dropout mechanism is applied to edges in the graph $G_{\mathcal{T}}$. After graph initialization, the unidirectional edges pointing from observed values to the missing ones are randomly dropped. Notably, the dropping rate is higher if the observed point is closer to the missing one, defined as: 
\begin{equation}
\label{equ:dropout}
    p(\Delta t) = e^{-\alpha \Delta t + \beta},
\end{equation}
where $\Delta t$ is the temporal distance, and $\alpha$, $\beta$ are hyperparameters controlling the maximum rate and decay speed respectively. Notably, as the average dropout rate is set to $p=0.1$ in PAST, the parameter $\beta$ is completely determined by $\alpha$ if $p$ is fixed, satisfying:
\begin{equation}
\beta = \log{\frac{pL^2}{\sum_{i=1}^L\sum_{j=1}^Le^{-\alpha |j-i|}}}
\end{equation}
where $L$ is the input sequence length, so that the number of dropout edges can be controlled.

This design is derived from the empirical observation in traffic time series that adjacent data points exhibit stronger dependencies, but stronger dependencies also indicate instability in incomplete sequences. For example, if a missing point mainly relies on its adjacent points, once the adjacent points are also missing in fiber or block missing scenarios, the model will be difficult to provide an accurate imputation result.
Therefore, by preferentially dropping proximal edges, the model is compelled to learn broader patterns, reducing local overfitting. This regularization is theoretically justified by analogy to dropout in neural networks, where selective edge removal promotes robustness \cite{dropout}.

\paragraph{External Info Injection.} As mentioned in Section \ref{sec:idea}, each graph-integrated layer receives a hidden vector from the corresponding cross-gated layer. The vector is added to the temporal graph $G_\mathcal{T}$ as an additional vertex, and points unidirectionally to all the other ve t, injecting information collected from spatio-temporal external features without reciprocal influence.

\paragraph{Feed Forward Process}
The adjacency matrix of $G_{\mathcal{T}}$ is defined as $\mathbf{A}_{\mathcal{T}} \in \mathbb{R}^{(L+1) \times (L+1)}$ ($L$ data-point vertices and one embedding vertex), with entries set to learnable parameters for existing edges and zero otherwise. The temporal layer applies a graph convolution operation to the vertex vectors $\mathbf{H} \in \mathbb{R}^{(L+1)\times d}$:
\begin{equation}
\begin{aligned}
\mathbf{H}^{(i)'} &= \mathbf{D}^{-1}_\mathcal{T}\mathbf{A}_{\mathcal{T}}\mathbf{H}^{(i)},\\
\mathbf{H}^{(i)''} &= \sigma(\mathbf{W}_{\mathcal{T}}\mathbf{H}^{(i)'T} + \mathbf{b}_\mathcal{T})^T,
\end{aligned}
\end{equation}
where $\bm{D}_\mathcal{T}\in \mathbb{R}^{(L+1) \times (L+1)}$ is the degree matrix of $\mathbf{A}_\mathcal{T}$, $\mathbf{W}_\mathcal{T}\in \mathbb{R}^{d \times d}$ and $\mathbf{b}\in \mathbb{R}^{d}$ are learnable parameters of a linear layer, and $\sigma$ is a non-linear activation function (e.g., ReLU).

\subsubsection{Spatial Layer}
The spatial layer models dependencies across nodes using a graph $G_\mathcal{S} = <V_\mathcal{S},E_\mathcal{S}>$ based on traffic network topology adjacency matrix $\mathbf{A}_\mathcal{S}$. 

\paragraph{Spatial Graph Building.} The graph is built directly from $\mathbf{A}_\mathcal{S}$, with vertices representing node embeddings from the temporal output. $\mathbf{A}_\mathcal{S}$ is determined by the distance between nodes:
\begin{equation}
    a_{ij}=\left\{
    \begin{aligned}
        &e^{-d^2_{ij}/\sigma^2},\quad\text{node $i,j$ are connected},\\
        &0,\quad\text{otherwise},\\
    \end{aligned}
    \right.
\end{equation}
where $d_{ij}$ is the distance between two nodes, and $\sigma$ is the standard deviation of $\{d_{ij}\}$.
Edge dropout is not required in this step, because the strength of spatial dependencies is determined by adjacency matrix $\mathbf{A}_\mathcal{S}$, which is not learnable. This design prevents the model from relying solely on adjacent nodes to learn spatial dependencies, thereby mitigating performance degradation.

\paragraph{Multi-order Convolution.} To perceive larger spatial spans and capture multi-order dependencies, the adjacency matrix is raised to powers from 0 to $K$:
$\mathbf{I}, \mathbf{A}_\mathcal{S}, \mathbf{A}^2_\mathcal{S}, \cdots, \mathbf{A}^K_\mathcal{S}$. For each order $k$, a graph convolution is performed on the features $\mathbf{H}^{(i)''}$ from the temporal layer: 
\begin{equation}
\mathbf{H}_k^{(i)'''} = \mathbf{D}_k^{-1/2} \mathbf{A}^k_\mathcal{S} \mathbf{D}_k^{-1/2} \mathbf{H}^{(i)''}, 
\end{equation}
where $\mathbf{D}_k$ is the degree matrix for $\mathbf{A}^k_\mathcal{S}$. The resulting $K+1$ hidden states $\{\mathbf{H}_k^{(i)'''}|k=0,1,\cdots,K\}$ are then fused through a linear layer:
\begin{equation}
\mathbf{H}^{(i+1)} = \sigma(\mathbf{W}_\mathcal{S} \left[\mathbf{H}_0, \cdots, \mathbf{H}_K \right] + \mathbf{b}_\mathcal{S}).
\end{equation}
This multi-order approach, inspired by higher-order graph convolutions \cite{GRIN}, captures broader spatial dependencies to enhance the model's ability to handle block missing scenarios.

\subsection{Cross-Gated Module}
\begin{figure}[t]
    \centering
    \includegraphics[width=\linewidth]{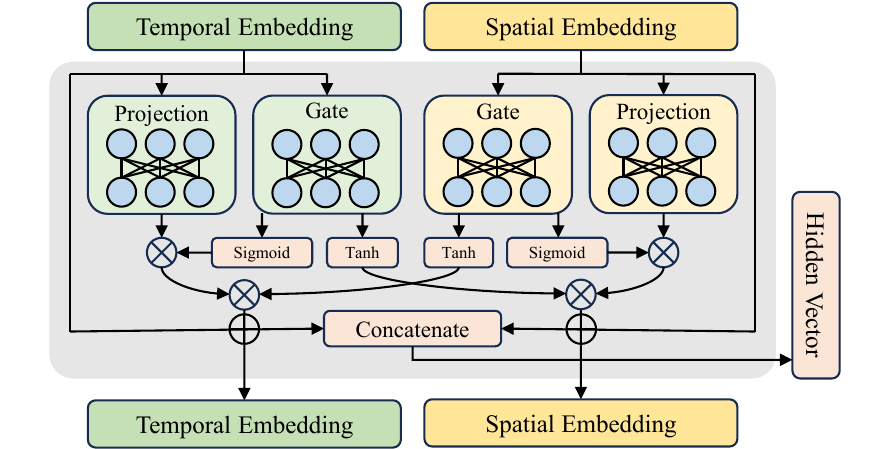}
    \caption{Cross-gated layer for auxiliary patterns}
    \Description{}
    \label{fig:cross_gated_unit}
    
\end{figure}

The cross-gated module (CGM) is designed to extract auxiliary spatio-temporal patterns from external information. It complements GIM by providing robust support especially for fiber and block missing scenarios. The core component of CGM is the cross-gated layer, whose detailed architecture is illustrated in Figure \ref{fig:cross_gated_unit}.

The input of the $i^{th}$ layer includes a spatial embedding vector $\bm{v_s} \in \mathbb{R}^d$ and a temporal embedding vector $\bm{v_t} \in \mathbb{R}^d$. An intuitive approach is to concatenate the two vectors and make projection through a $2d\times 2d$ linear layer and an activation function. However, this approach directly blends the spatial and temporal patterns without explicitly modeling the detailed interactions, thereby limiting the fitting ability.

Inspired by the gated linear unit (GLU) \cite{glu}, which filters information via linear and sigmoid paths, we introduce a cross-gated mechanism for bidirectional spatio-temporal gating, which contains two main steps: feature projection and pattern interactions.

\subsubsection{Feature Projection}
Instead of introducing a $2d\times 2d$ linear layer, the cross-gated layer contains four $d \times d$ sub-layers, two for the temporal embedding vector and two for the spatial one. For the spatial/temporal vector, one sub-layer is for projection and the other is for interaction. The projection layer extracts features from vectors, while the gated layer determines the interaction details.
\begin{equation}
    \begin{aligned}
\bm{v_{sp}} = \bm{W_{sp}v_s},& \quad\bm{v_{tp}} = \bm{W_{tp}v_t},\\
\bm{v_{sg}} = \bm{W_{sg}v_s},& \quad\bm{v_{tg}} = \bm{W_{tg}v_t}
    \end{aligned}
\end{equation}
where $\bm{v_{sp}}$ and $\bm{v_{tp}}$ denote the outputs of spatial/temporal projection layers, while $\bm{v_{sg}}$ and $\bm{v_{tg}}$ denote the outputs of the gated layers.

\subsubsection{Feature Selection and Interaction}
The temporal gated vector $\bm{v_{tg}}$ can be used to select relevant features in the temporal projection vector $\bm{v}_{tp}$, and suppress irrelevant ones. This is realized via the sigmoid activation function and element-wise multiplication. Symmetrically, the spatial gated vector also filters the spatial projection vector. Additionally, in order to accommodate spatio-temporal interactions, the gated vectors are refined by the tanh function and mutually multiplied with the projection vectors from the opposite domain, because the output range of tanh $(-1, 1)$ can simulate both positive and negative relationships between spatial and temporal patterns. In summary, the feature selection and interaction process is formalized as follows:
\begin{equation}
\begin{aligned}
    \bm{v_{sp}} &\leftarrow \bm{v_{sp}} \cdot Sigmoid(\bm{v_{sg}}) \cdot Tanh(\bm{v}_{tg})\\
    \bm{v_{tp}} &\leftarrow \bm{v_{tp}} \cdot Sigmoid(\bm{v_{tg}}) \cdot Tanh(\bm{v}_{sg}).
\end{aligned}
\end{equation}

Finally, the residual connection is applied to mitigate gradient vanishing, 
\begin{equation}
\begin{aligned}
\bm{v_s}' &= \bm{v_s} + \bm{v_{sp}},\\
\bm{v_t}' &= \bm{v_t} + \bm{v_{tp}}.
\end{aligned}
\end{equation}

The refined vectors are output to the subsequent layer, and concatenated together to form the hidden vector, which encapsulates the extracted auxiliary patterns and is forwarded to the corresponding graph-integrated layer for inter-module information exchange.

In summary, the cross-gated layer enhances spatio-temporal representation capacity while maintaining efficiency compared to a standard linear layer. As the embedding dimension is $d$, the input and output dimensions of normal linear layers are $2d$, indicating that the number of parameters is $4d^2$. As shown in Figure \ref{fig:cross_gated_unit}, the cross-gated layer is composed of four $d\times d$ linear layers, so the total number of parameters is also $4d^2$. However, the layer concretizes dependency learning through self- and cross-representations, excelling in the dual-pattern scenarios like traffic imputation.

\subsection{Overall Analysis}
Above all, PAST introduces several key innovations to address traffic time series imputation across diverse missing types. First, it disentangles spatio-temporal patterns into primary patterns, derived from internal data relationships, and auxiliary patterns, inferred from external features, enabling targeted extraction that aligns with specific missing scenarios. Second, the dual-module architecture, comprising GIM and CGM, facilitates independent yet interactive processing: GIM employs dynamic directed graphs with interval-aware dropout and multi-order convolutions to robustly capture primary patterns despite varying missing positions, while CGM utilizes a cross-gated mechanism for bidirectional spatio-temporal interactions, enhancing representation of long-range dependencies. Third, inter-module synergy is achieved through hidden vector exchanges and an ensemble training framework inspired by residual fitting, positioning CGM as an enhancer for patterns overlooked by GIM. These advancements collectively promote generalization, stability, and accuracy by mitigating overfitting and adapting to heterogeneous missing conditions.

The computational complexity of PAST is analyzed as follows. Let $n$ denote the number of layers, $T$ the time steps, $N$ the nodes, and $d$ the embedding dimension. The CGM's embedding and cross-gated operations scale as $O(nTNd^2)$, dominated by linear projections and gating. In the GIM, the temporal layer constructs per-node graphs with $T+1$ nodes and up to $O(T^2)$ edges (in fully observed cases due to bidirectional connections among observed nodes), yielding $O(nT^2Nd^2)$ for CGM propagation. The spatial layer, using the topology adjacency matrix $A_{\mathcal{S}}$ expanded to larger spans via powering, adds $ O(nTNKd^2) $. Thus, the overall time complexity is $ O(nTNd^2(1+T+K)) $, primarily governed by the temporal layers in GIM for long sequences.

\section{Experiments}
\subsection{Basic Information}
\subsubsection{Datasets}
Experiments are conducted on three real-world datasets: METR-LA \cite{stgcn}, PeMS-Bay \cite{dcrnn}, and LargeST-SD \cite{largest}.
\begin{itemize}
    \item \textbf{METR-LA}. This dataset contains traffic speeds collected from 207 loop detectors on highways in Los Angeles, aggregated into 5-minute intervals over a period of four months (March 1 to June 30, 2012). It captures spatio-temporal patterns in urban traffic flow, with a total of 34,272 time steps.
    \item \textbf{PeMS-Bay}. This dataset includes traffic speed readings from 325 sensors in the San Francisco Bay Area, spanning six months (January 1 to May 31, 2017) at 5-minute intervals, resulting in 52,116 time steps.
    \item \textbf{LargeST-SD}. LargeST is a large-scale benchmark dataset for traffic forecasting, comprising five years (2017-2021) traffic flow data from 8,600 sensors across California highways. SD (San Diego) is one of its subsets that contains 716 sensors and 1739 edges.
\end{itemize}
All the three datasets are down-sampled via window averaging to a 15-minute time interval.

\subsubsection{Baselines}
PAST is compared with seven imputation models, including four classical models (KNN \cite{KNN}, linear interpolation, MICE \cite{mice} and TIDER \cite{TIDER}), two deep learning models for general imputation (BRITS \cite{BRITS} and GRIN \cite{GRIN}) and one traffic-specific imputation model (STCPA \cite{STCPA}).

\begin{itemize}
    \item \textbf{KNN}\cite{KNN}. The K-Nearest Neighbors method fills missing values based on the average of the k most similar observed instances.
    \item \textbf{Linear}. Linear interpolation estimates missing values by connecting adjacent observed points with a straight line.
    \item \textbf{MICE}\cite{mice}. Multiple Imputation by Chained Equations is an iterative regression-based method that models each variable conditionally on others.
    \item \textbf{TIDER}\cite{TIDER}. This is a matrix factorization-based model that decomposes temporal patterns into trend, seasonality, and local bias components for multivariate imputation.
    \item \textbf{BRITS}\cite{BRITS}. Bidirectional Recurrent Imputation for Time Series is an RNN-based model that imputes missing values by learning bidirectional temporal dependencies.
    \item \textbf{GRIN}\cite{GRIN}. Graph Recurrent Imputation Network integrates GNNs with RNNs to capture spatial and temporal relationships in incomplete time series.
    \item \textbf{STCPA}\cite{STCPA}. STCPA is a traffic-specific model with cyclic perceptron training framework and self-attention to handle multiple missing types.
\end{itemize}

\subsubsection{Environments}
The model is implemented in PyTorch and trained on a single NVIDIA A6000 GPU with 48GB memory, running on Ubuntu 20.04.

\subsection{Settings}
All models are trained using a self-supervised framework with mean squared error (MSE) loss on observed values. In order to simulate missing data conditions, a part of values are masked. Experiments cover 9 missing conditions across the three datasets:
\begin{itemize}
\item Random missing: Missing rate $r$ is set to $0.2$, $0.4$, and $0.6$, where values are randomly masked independently.  
\item Fiber missing: The missing rate is fixed to $r=0.4$, with consecutive missing lengths. The length of each missing segment follows uniform distribution, with the maximum missing length $l = 24, 48, 96$ (representing 6, 12, and 24 hours at 15-minute intervals).  
\item Block missing: The missing rate is fixed to $r=0.4$, with consecutive missing lengths and spans. The maximum lengths are set to $48$ and $96$, and the maximum spans are set to $5$ and $10$. For example, $(48, 5)$ means that the missing length follows a uniform distribution from 1 to 48, and each missing block contains five connected nodes in the graph.
\end{itemize}

Subsequently, each dataset is split into training (80\%), and test (20\%) sets. For evaluation, the imputation of missing values in the training set is called offline / in-sample condition, which simulates the preprocessing of a dataset before further analysis. The imputation in the test set is called online / out-of-sample condition, where a trained model is directly used for the imputation of the future new sequences \cite{GRIN}. The imputation accuracy is reported by Root Mean Squared Error (RMSE) and Mean Absolute Error (MAE).

\begin{table}[t]
\caption{Hyper-parameter Settings}
\label{tab:params}
\centering
\begin{tabular}{ccc}\hline
Parameter&Symbol&Value\\\hline
Learning Rate&$lr$&$10^{-4}$\\
Batch Size&-&32\\
Optimizer&-&Adam\\
Dropout Rate&$p$&0.1\\
Sequence Length&$L$&96\\\hline
Layer Num&$n$&3\\
Embedding Dim&$d$&64\\
Dropout Param&$\alpha$&0.1\\
Multi-hop Order&$K$&2\\\hline
\end{tabular} 
\end{table}

For PAST, key hyperparameters are listed in Table \ref{tab:params}. Baselines use their default or optimized hyperparameters from the original implementations.

\subsection{Accuracy Evaluation}
\subsubsection{Random Missing}
\begin{table*}[t]
\centering
\caption{Imputation Accuracy on Random Missing Conditions}
\label{tab:exp_random}
\begin{tabular}{cc|cccccccc|cccc}\hline
\multirow{2}*{$r$}&\multirow{2}*{Dataset}&\multicolumn{8}{c|}{Offline / In-sample}&\multicolumn{4}{c}{Online / Out-of-sample}\\
~&~&KNN&Linear&MICE&TIDER&BRITS&GRIN&STCPA&\textbf{PAST}&BRITS&GRIN&STCPA&\textbf{PAST}\\\hline
\multicolumn{14}{c}{\textbf{RMSE}}\\\hline

\multirow{3}*{0.2}&METR&0.7187&0.2860&0.2913&1.1390&0.3109&\underline{0.2218}&0.2915&\textbf{0.1952}&0.4219&\underline{0.2191}&0.3561&\textbf{0.1979}\\
~&PeMS&0.9022&0.2139&0.2385&0.8363&0.2898&\underline{0.1900}&0.3037&\textbf{0.1897}&0.3671&\underline{0.2069}&0.3876&\textbf{0.2030}\\
~&LargeST&0.7828&0.1075&0.0978&0.8812&0.1880&\underline{0.0906}&0.1536&\textbf{0.0769}&0.2244&\underline{0.0869}&0.1658&\textbf{0.0767}\\\hline

\multirow{3}*{0.4}&METR&0.7342&0.3167&0.3459&1.1140&0.3281&\underline{0.2397}&0.3030&
\textbf{0.2082}&0.4475&\underline{0.2419}&0.3965&\textbf{0.2108}\\
~&PeMS&0.9212&0.2551&0.3329&1.1350&0.3204&\underline{0.2224}&0.3254&\textbf{0.1996}&0.3978&\underline{0.2476}&0.4064&\textbf{0.2104}\\
~&LargeST&0.7858&0.1173&0.1104&0.8792&0.2301&\underline{0.0952}&0.1520&\textbf{0.0914}&0.2558&\underline{0.0931}&0.2528&\textbf{0.0916}\\\hline

\multirow{3}*{0.6}&METR&0.7646&0.3665&0.4223&0.9362&0.3540&\textbf{0.2754}&0.3396&\underline{0.2820}&0.5011&\underline{0.2903}&0.4240&\textbf{0.2888}\\
~&PeMS&0.9619&0.3259&0.4505&1.0890&0.3697&\underline{0.2946}&0.3569&\textbf{0.2564}&0.4375&\underline{0.3177}&0.4429&\textbf{0.2672}\\
~&LargeST&0.7949&0.1378&0.1316&0.8726&0.2200&\textbf{0.1069}&0.1543&\textbf{0.0978}&0.2549&\underline{0.1064}&0.1862&\textbf{0.0986}\\\hline
\multicolumn{14}{c}{\textbf{MAE}}\\\hline

\multirow{3}*{0.2}&METR&0.4520&\underline{0.1284}&0.1575&0.8104&0.1316&0.1331&0.1390&\textbf{0.1007}&0.2172&\underline{0.1314}&0.1766&\textbf{0.1038}\\
~&PeMS&0.5274&0.1025&0.1297&0.5253&0.1354&\textbf{0.0977}&0.1328&\underline{0.0981}&0.1827&\underline{0.1015}&0.1762&\textbf{0.1008}\\
~&LargeST&0.5729&0.0697&0.0594&0.7271&0.1205&\underline{0.0590}&0.0790&\textbf{0.0488}&0.1365&\underline{0.0557}&0.1013&\textbf{0.0479}\\\hline

\multirow{3}*{0.4}&METR&0.4612&0.1423&0.1945&0.7981&0.1379&\underline{0.1280}&0.1358&\textbf{0.1018}&0.2317&\underline{0.1305}&0.1805&\textbf{0.1036}\\
~&PeMS&0.5376&0.1180&0.1752&0.7159&0.1485&\underline{0.1108}&0.1436&\textbf{0.1002}&0.1961&\underline{0.1166}&0.1810&\textbf{0.1035}\\
~&LargeST&0.5748&0.0750&0.0731&0.7210&0.1481&\underline{0.0656}&0.0845&\textbf{0.0591}&0.1612&\underline{0.0633}&0.1257&\textbf{0.0583}\\\hline

\multirow{3}*{0.6}&METR&0.4768&0.1653&0.2435&0.6769&0.1495&\textbf{0.1425}&0.1503&\underline{0.1475}&0.2570&\textbf{0.1517}&0.1890&\underline{0.1534}\\
~&PeMS&0.5571&\underline{0.1446}&0.2353&0.6728&0.1701&0.1554&0.1596&\textbf{0.1261}&0.2120&\underline{0.1599}&0.2011&\textbf{0.1299}\\
~&LargeST&0.5812&0.0857&0.0831&0.7172&0.1451&\underline{0.0726}&0.0838&\textbf{0.0631}&0.1608&\underline{0.0711}&0.1432&\textbf{0.0627}\\\hline
\end{tabular}
\end{table*}

Table \ref{tab:exp_random} shows the experiment results in random missing scenarios, where three main phenomena can be observed. First, PAST exhibits the best performance in most cases, with an average reduction of approximately 8.2\% in RMSE and 11.2\% in MAE compared to the strongest baseline, which proves its generality in random missing scenarios. Second, deep learning models generally achieve superior results, due to their ability to learn complex, non-linear relationships from incomplete data. GRIN outperforms the others thanks to its dual-dimensional approach that simultaneously addresses spatial and temporal modeling. Third, traditional models such as KNN and matrix factorization, exhibit lower effectiveness overall because they rely on simpler assumptions like nearest-neighbor similarity or low-rank approximations, which struggle with heterogeneous patterns; however, linear interpolation and MICE do not lag far behind. This relatively narrow performance gap underscores the simplicity of dependencies in random missing scenarios, where most underlying patterns can be inferred from adjacent observed data points alone.

\subsubsection{Fiber Missing}
\begin{table*}[t]
\centering
\caption{Imputation Accuracy on Fiber Missing Conditions with $r=0.4$}
\label{tab:exp_fiber}
\begin{tabular}{cc|cccccccc|cccc}\hline
\multirow{2}*{$l$}&\multirow{2}*{Dataset}&\multicolumn{8}{c|}{Offline / In-sample}&\multicolumn{4}{c}{Online / Out-of-sample}\\
~&~&KNN&Linear&MICE&TIDER&BRITS&GRIN&STCPA&\textbf{PAST}&BRITS&GRIN&STCPA&\textbf{PAST}\\\hline
\multicolumn{14}{c}{\textbf{RMSE}}\\\hline

\multirow{3}*{24}&METR&0.7317&0.591&0.5832&1.0400&0.3546&\underline{0.3282}&0.4283&\textbf{0.2867}&0.5449&\underline{0.3739}&0.5457&\textbf{0.3083}\\
~&PeMS&0.9184&0.6739&0.6683&1.0140&\underline{0.3449}&0.3504&0.4347&\textbf{0.3098}&0.4304&\underline{0.3935}&0.5447&\textbf{0.3461}\\
~&LargeST&0.7851&0.3123&0.1746&0.8820&0.1981&\underline{0.1299}&0.1855&\textbf{0.1196}&0.2395&\underline{0.1391}&0.2275&\textbf{0.1282}\\\hline

\multirow{3}*{48}&METR&0.7320&0.7201&0.6866&1.2060&\underline{0.3558}&0.3919&0.4641&\textbf{0.3009}&0.6054&\underline{0.444}&0.6023&\textbf{0.3440}\\
~&PeMS&0.9180&0.9008&0.8305&0.8205&\underline{0.3607}&0.3702&0.4731&\textbf{0.3179}&0.4386&\underline{0.4355}&0.5719&\textbf{0.3589}\\
~&LargeST&0.7867&0.5509&0.2630&0.8639&0.2284&\underline{0.1579}&0.1925&\textbf{0.1193}&0.2688&\underline{0.1782}&0.2437&\textbf{0.1283}\\\hline
\multirow{3}*{96}&METR&0.7371&0.8584&0.822&1.0590&\underline{0.3832}&0.5006&0.4514&\textbf{0.3391}&0.6129&\underline{0.5224}&0.5457&\textbf{0.3870}\\
~&PeMS&0.9233&1.0981&0.9730&0.8674&\underline{0.3816}&0.4013&0.4583&\textbf{0.3501}&0.4625&\underline{0.4621}&0.5887&\textbf{0.3880}\\
~&LargeST&0.7844&0.9371&0.5696&0.8638&0.1878&\underline{0.1638}&0.1831&\textbf{0.1430}&0.2316&\underline{0.1974}&0.2347&\textbf{0.1648}\\\hline

\multicolumn{14}{c}{\textbf{MAE}}\\\hline

\multirow{3}*{24}&METR&0.4585&0.2926&0.3117&0.7460&\underline{0.1472}&0.1604&0.1703&\textbf{0.1408}&0.2486&\underline{0.1819}&0.3202&\textbf{0.1506}\\
~&PeMS&0.5365&0.3073&0.3218&0.6232&\underline{0.1595}&0.1692&0.1882&\textbf{0.1579}&0.2083&\underline{0.1844}&0.2278&\textbf{0.1696}\\
~&LargeST&0.5743&0.1935&0.1060&0.7262&0.1263&\underline{0.0867}&0.1131&\textbf{0.0753}&0.1464&\underline{0.0896}&0.1426&\textbf{0.0806}\\\hline

\multirow{3}*{48}&METR&0.4598&0.3850&0.366&0.8525&\underline{0.1507}&0.2011&0.1677&\textbf{0.1466}&0.2942&\underline{0.2295}&0.3169&\textbf{0.1635}\\
~&PeMS&0.5350&0.4338&0.3908&0.5214&\underline{0.1755}&0.1837&0.1817&\textbf{0.1640}&0.2146&\underline{0.2085}&0.2346&\textbf{0.1774}\\
~&LargeST&0.5754&0.3593&0.1498&0.7063&0.1465&\underline{0.1098}&0.1133&\textbf{0.0745}&0.1666&\underline{0.1191}&0.1411&\textbf{0.0915}\\\hline

\multirow{3}*{96}&METR&0.4626&0.4830&0.4415&0.7711&\underline{0.1711}&0.2924&0.1770&\textbf{0.1607}&0.3116&\underline{0.3104}&0.3272&\textbf{0.1890}\\
~&PeMS&0.5383&0.5623&0.4700&0.5470&\underline{0.1734}&0.2107&0.1907&\textbf{0.1606}&\underline{0.2247}&0.2376&0.2491&\textbf{0.1942}\\
~&LargeST&0.5737&0.6541&0.3105&0.7085&0.1208&\underline{0.1044}&0.1152&\textbf{0.0859}&0.1411&\underline{0.1184}&0.1588&\textbf{0.0951}\\\hline
\end{tabular}
\end{table*}

Table \ref{tab:exp_fiber} shows the results in fiber missing scenarios, where three key points should be concentrated on. 
First, PAST achieves the best performance across all evaluated settings, with more pronounced improvements compared to random missing: in offline (in-sample) evaluations, it reduces RMSE by approximately 12.8\% and MAE by 10.1\%, while in online (out-of-sample) evaluations, the reductions reach 18.2\% in RMSE and 19.4\% in MAE. This substantial advantage highlights the efficacy of PAST's primary-auxiliary paradigm for fiber missing, where extended temporal gaps necessitate capturing long-term periodicity driven by external factors. 
Second, compared to random missing, traditional models experience a marked decline in accuracy. This is because their simplistic assumptions falter when prolonged gaps disrupt local relationships, rendering them inadequate for the more complex patterns. In contrast, deep learning models exhibit heightened advantages by better leveraging bidirectional RNNs and attention mechanisms to infer latent temporal structures from sparse observations. Third, the baseline models show significant performance degradation from offline to online scenarios, whereas PAST maintains stable accuracy, underscoring its superior generalization capabilities. This indicates that existing models may overfit to the training data, while the primary-auxiliary framework of PAST ensures precise capture of overlooked patterns without introducing bias.

\subsubsection{Block Missing}
\begin{table*}[t]
\centering
\caption{Imputation Accuracy on Block Missing Conditions with $r=0.4$}
\label{tab:exp_block}
\begin{tabular}{cc|cccccccc|cccc}\hline
\multirow{2}*{$(l,s)$}&\multirow{2}*{Dataset}&\multicolumn{8}{c|}{Offline / In-sample}&\multicolumn{4}{c}{Online / Out-of-sample}\\
~&~&KNN&Linear&MICE&TIDER&BRITS&GRIN&STCPA&\textbf{PAST}&BRITS&GRIN&STCPA&\textbf{PAST}\\\hline
\multicolumn{14}{c}{\textbf{RMSE}}\\\hline

\multirow{3}*{(48, 5)}&METR&0.7363&0.6969&0.6775&1.0820&\underline{0.3826}&0.4015 
&0.4357&\textbf{0.3114}&0.5856&\underline{0.4391}&0.6020&\textbf{0.3454}\\
~&PeMS&0.9257&0.8759&0.8384&1.0240&\underline{0.3778}&0.3808&0.4421&\textbf{0.3247}&0.4500&\underline{0.4464}&0.5332&\textbf{0.3595}\\
~&LargeST&0.8070&0.5445&0.2497&0.9059&0.2244&\underline{0.1478}&0.1875&\textbf{0.1348}&0.2510&\underline{0.1517}&0.2462&\textbf{0.1433}\\\hline

\multirow{3}*{(48, 10)}&METR&0.7346&0.6883&0.6662&1.0920&\underline{0.3825}&0.6804&0.4707&\textbf{0.3090}&\underline{0.5459}&0.6725&0.6591&\textbf{0.3442}\\
~&PeMS&0.9240&0.8418&0.8096&0.8674&\underline{0.3682}&0.3700&0.4813&\textbf{0.3220}&0.4614&\underline{0.4502}&0.5588&\textbf{0.3634}\\
~&LargeST&0.8052&0.5286&0.2473&0.9447&0.2106&\underline{0.1455}&0.1952&\textbf{0.1306}&0.2393&\underline{0.1590}&0.2488&\textbf{0.1404}\\\hline

\multirow{3}*{(96, 5)}&METR&0.7348&0.8343&0.7898&1.0210&\underline{0.3896}&0.4394&0.4475&\textbf{0.3414}&0.5444&\underline{0.4808}&0.6395&\textbf{0.3590}\\
~&PeMS&0.9211&1.1007&0.9337&0.9770&\underline{0.3859}&0.4058&0.4630&\textbf{0.3412}&\underline{0.4645}&0.4702&0.5531&\textbf{0.3792}\\
~&LargeST&0.8074&0.9240&0.4380&0.9004&0.2811&\underline{0.1766}&0.1851&\textbf{0.1382}&0.3282&\underline{0.1891}&0.2412&\textbf{0.1493}\\\hline

\multicolumn{14}{c}{\textbf{MAE}}\\\hline

\multirow{3}*{(48, 5)}&METR&0.4602&0.3671&0.3716&0.7723&\underline{0.1609}&0.2160&0.1792&\textbf{0.1479}&0.2825&\underline{0.2378}&0.3120&\textbf{0.1575}\\
~&PeMS&0.5394&0.4182&0.3987&0.6356&\underline{0.1735}&0.1927&0.1955&\textbf{0.1632}&0.2212&\underline{0.2193}&0.2529&\textbf{0.1759}\\
~&LargeST&0.5910&0.3554&0.1495&0.7338&0.1407&\underline{0.0944}&0.1152&\textbf{0.0843}&0.1549&\underline{0.1008}&0.1512&\textbf{0.0888}\\\hline

\multirow{3}*{(48, 10)}&METR&0.4617&0.3590&0.3617&0.7727&\underline{0.1589}&0.3403&0.1711&\textbf{0.1482}&\underline{0.2555}&0.3422&0.3272&\textbf{0.1647}\\
~&PeMS&0.5380&0.3967&0.3830&0.5825&\underline{0.1723}&0.1808&0.1839&\textbf{0.1623}&0.2248&\underline{0.2091}&0.2448&\textbf{0.1761}\\
~&LargeST&0.5905&0.3408&0.1469&0.7770&0.1391&\underline{0.1048}&0.1145&\textbf{0.0850}&0.1546&\underline{0.1112}&0.1440&\textbf{0.0893}\\\hline

\multirow{3}*{(96, 5)}&METR&0.4603&0.4621&0.4331&0.9567&0.1878&0.2461&\underline{0.1806}&\textbf{0.1689}&\underline{0.2607}&0.2726&0.3505&\textbf{0.1790}\\
~&PeMS&0.5371&0.5614&0.4559&0.6106&\underline{0.1780}&0.2084&0.1932&\textbf{0.1731}&\underline{0.2274}&0.2373&0.2557&\textbf{0.1877}\\
~&LargeST&0.5923&0.6438&0.2425&0.7358&0.1431&0.1188&\underline{0.1164}&\textbf{0.0905}&0.1567&\underline{0.1248}&0.1654&\textbf{0.0961}\\\hline
\end{tabular}
\end{table*}

Table \ref{tab:exp_block} records the RMSE and MAE in spatio-temporal contiguous gaps at $r=0.4$ with different $(l,s)$ pairs, including $(48, 5)$, $(48, 10)$, and $(96, 5)$. In this three conditions, PAST's advantages become even more pronounced, consistently achieving the highest imputation accuracy across all experimental settings: in offline evaluations, it reduces RMSE by approximately 14.4\% and MAE by 10.9\%, while in online evaluations, the improvements escalate to 26.2\% in RMSE and 31.6\% in MAE. This heightened superiority stems from PAST's tailored handling of spatially and temporally contiguous gaps. 
Additionally, compared to fiber missing, GRIN experiences a more substantial accuracy decline in block missing, as its GNN architecture makes itself particularly vulnerable when block missing disrupts adjacent nodes' topological connectivity. In contrast, other models like BRITS or STCPA, which emphasize temporal patterns, suffer less pronounced impacts, allowing them to fall back on long-term temporal cues with minimal additional degradation. 
The amplified gap between offline and online performance for baselines further illustrates PAST's robustness. Its ensemble training framework and bidirectional module interactions prevent overfitting to training-specific spatial artifacts and generalize to unseen block configurations.

\subsubsection{Summary}
In summary, the experimental results across three datasets and 27 diverse missing conditions demonstrate the superior efficacy of PAST in traffic time series imputation, outperforming seven state-of-the-art baselines by 14.7\% in RMSE and 15.8\% in MAE on average.
The advantage is most pronounced in structured missing scenarios, by up to 26.2\% in RMSE and 31.6\% in MAE.
This highlights the robustness to extended temporal gaps and large-scale spatial voids. This effectiveness follows from a primary–auxiliary disentanglement that aligns pattern extraction with missing types. The graph-integrated module (GIM) captures primary patterns using dynamic directed graphs, interval-aware dropout to reduce overfitting, and multi-hop convolutions for topological propagation. The cross-gated module (CGM) complements GIM by modeling auxiliary patterns through bidirectional gating on timestamps and node features, enabling efficient inference of long-term periodicity and spatial similarity.

These innovations not only address the limitations of prior frameworks, but also forge a link between external-driven patterns and imputation stability. To further dissect the contributions of these architectural elements and validate their necessity, we conduct an ablation study in the following section.

\subsection{Ablation Studies}
\begin{table*}[t]
\centering
\caption{Ablation studies on Cross-gated Unit and Graph-Integrated Unit with missing rate $r=0.4$.}
\label{tab:ablation}
\begin{tabular}{cc|ccccc|ccccc}\hline
\multirow{2}*{$(l,s)$}&\multirow{2}*{Dataset}&\multicolumn{5}{c|}{Offline / In-sample}&\multicolumn{5}{c}{Online / Out-of-sample}\\
~&~&\textbf{PAST}&w/o CGM&w/o GIM&CGL-LL&GNN-RNN&\textbf{PAST}&w/o CGM&w/o GIM&CGL-LL&GNN-RNN\\\hline
\multicolumn{12}{c}{\textbf{RMSE}}\\\hline
\multirow{3}*{(1, 1)}&METR&\textbf{0.2082}&0.2687&0.8131&0.2700&0.2387&\textbf{0.2108}&0.2622&1.0633&0.2609&0.2628\\
~&PeMS&\textbf{0.1996}&0.2520&0.4885&0.2129&0.2280&\textbf{0.2104}&0.2544&0.5382&0.2216&0.2451\\
~&LargeST&\textbf{0.0914}&0.1513&0.2185&0.1094&0.1012&\textbf{0.0916}&0.1319&0.3265&0.0996&0.1154\\\hline

\multirow{3}*{(48, 1)}&METR&\textbf{0.3009}&0.4022&0.8523&0.3849&0.3376&\textbf{0.3440}&0.3854&1.1021&0.3644&0.4459\\
~&PeMS&\textbf{0.3179}&0.5047&0.5024&0.3428&0.3501&\textbf{0.3589}&0.5219&0.5385&0.3762&0.3927\\
~&LargeST&\textbf{0.1193}&0.1916&0.2195&0.1353&0.1440&\textbf{0.1283}&0.1922&0.3218&0.1434&0.1625\\\hline

\multirow{3}*{(48, 5)}&METR&\textbf{0.3090}&0.4331&0.8494&0.4475&0.3420&\textbf{0.3442}&0.4043&1.1031&0.4072&0.4351\\
~&PeMS&\textbf{0.3220}&0.5053&0.4937&0.3355&0.3605&\textbf{0.3634}&0.5192&0.5266&0.3693&0.4045\\
~&LargeST&\textbf{0.1306}&0.2131&0.2137&0.1608&0.1348&\textbf{0.1404}&0.2085&0.3079&0.1687&0.1543\\\hline

\multicolumn{12}{c}{\textbf{MAE}}\\\hline
\multirow{3}*{(1, 1)}&METR&\textbf{0.1018}&0.1325&0.6611&0.1386&0.1252&\textbf{0.1036}&0.1183&1.1333&0.1222&0.1548\\
~&PeMS&\textbf{0.1002}&0.1246&0.2386&0.1079&0.1197&\textbf{0.1035}&0.1262&0.2897&0.1109&0.1244\\
~&LargeST&\textbf{0.0591}&0.0956&0.1467&0.0710&0.0728&\textbf{0.0583}&0.0964&0.2066&0.0748&0.0702\\\hline

\multirow{3}*{(48,1)}&METR&\textbf{0.1466}&0.2291&0.7265&0.2353&0.1629&\textbf{0.1635}&0.2052&1.2147&0.1874&0.2281\\
~&PeMS&\textbf{0.1640}&0.2927&0.2523&0.1969&0.1695&\textbf{0.1774}&0.2799&0.5422&0.1979&0.1948\\
~&LargeST&\textbf{0.0745}&0.1321&0.1482&0.0875&0.0919&\textbf{0.0915}&0.1506&0.2036&0.1045&0.1025\\\hline

\multirow{3}*{(48, 5)}&METR&\textbf{0.1482}&0.2305&0.7221&0.2593&0.1761&\textbf{0.1647}&0.2244&1.2169&0.2471&0.2181\\
~&PeMS&\textbf{0.1623}&0.2700&0.2438&0.1762&0.1839&\textbf{0.1761}&0.2714&0.2782&0.1877&0.2019\\
~&LargeST&\textbf{0.0850}&0.1540&0.1457&0.1125&0.0857&\textbf{0.0893}&0.1485&0.1948&0.1090&0.0959\\\hline
\end{tabular}
\end{table*}

To validate the necessity of the key components in PAST, we conduct ablation experiments on the METR-LA, PeMS-Bay, and LargeST-SD datasets under representative missing conditions. We evaluate PAST's variants by removing or modifying modules, and then report changes of RMSE and MAE. The experiment results are shown in Table \ref{tab:ablation}.

\subsubsection{Effect of Cross-Gated Module}
To evaluate the contributions of the Cross-Gated Module (CGM), we conduct ablation experiments from two perspectives: its role in leveraging auxiliary patterns and facilitating effective spatio-temporal interactions.

First, the CGM is removed from PAST, so imputation relies solely on the graph-integrated module (GIM) by learning primary patterns from internal data relationships. This variant results in a substantial degradation in performance: an average increase of 31.8\% in RMSE for offline evaluations and 15.6\% in RMSE for online evaluations. These declines underscore the critical supplementary role of auxiliary patterns.

Next, to assess the effectiveness of the cross-gated layers in modeling interactions between spatial and temporal auxiliary patterns, we replace them with linear layers followed by ReLU activations (denoted as CGL-LL in Table \ref{tab:ablation}), maintaining equivalent parameter counts for a fair comparison. As shown in the results, this substitution leads to an average RMSE increase of 15.8\% in online settings and 9.6\% in offline settings. This performance drop highlights the innovation of the bidirectional gating mechanism in CGM, which enables dynamic and non-linear fusion of embedded external features, thus preventing loss of representational power and enhancing robustness against varying missing positions.

Overall, these ablations confirm that CGM fulfills its auxiliary function in two key ways: by fully exploiting external information to address gaps in primary pattern learning, and by establishing an effective mechanism for spatio-temporal pattern extraction and interaction, which collectively elevate PAST's generalization and accuracy beyond GIM-centric approaches.


\subsubsection{Effect of Graph-Integrated Module}
To assess the contributions of the Graph-Integrated Module (GIM), we perform ablation experiments from two aspects, emphasizing its pivotal role as the backbone for capturing primary spatio-temporal patterns and its synergy within the primary-auxiliary framework.

First, the whole GIM is removed from PAST. The imputation completely depends on the cross-gated module (CGM) by modeling auxiliary patterns. This configuration yields a dramatic performance downturn: an average RMSE escalation of 52.9\% in offline evaluations and 58.8\% in online evaluations. These prove GIM's indispensable role in the architecture, as it is responsible for extracting core internal relationships. Without GIM, CGM's reliance on external cues alone proves insufficient for comprehensive pattern recovery, particularly in random missing where short-term dependencies prevail. Moreover, the consistent performance across varying missing ratios and extents demonstrates that CGM's imputation efficacy remains independent of missing lengths or spatial scopes.

Second, GIM's pure GNN architecture is substituted with a hybrid "RNN+GNN" structure (similar to GRIN), where temporal dependencies are processed via bidirectional RNNs. Accordingly, the hidden states transmitted from CGM are utilized as the RNN's initial hidden vectors to preserve module interactions. The results are recorded in the "GNN-RNN" column in Table \ref{tab:ablation}, indicating an RMSE rise of 10.6\% in offline scenarios and 16.4\% in online scenarios. This may because of RNNs' inherent challenges in modeling long-term dependencies amid continuous absences, as their sequential nature amplifies error propagation in fiber or block missing. In essence, the pure GNN design in GIM more effectively consolidates primary patterns by treating incomplete sequences as dynamic graphs. It avoids the limitations of recurrent dependencies, and is more robust against positional randomness.

In conclusion, these ablations validate GIM's role as the model's core backbone, whose architecture proficiently encapsulates spatio-temporal interdependencies among data points, addressing the overfitting risks and biased estimations. The complementary interplay of the primary-auxiliary paradigm empowers PAST to adeptly accommodate diverse missing conditions.

\subsection{Sensitivity Analysis}
In this section, we conduct experiments to evaluate the influence of some important hyperparameters, including the dropout parameter $\alpha$ in Equation (\ref{equ:dropout}), the spatial graph extension order $K$, and the embedding dimension $d$. Experiments are conducted on the METR-LA dataset in three missing scenarios: random missing with $r=0.4$, fiber missing with $l=48$, and block missing with $(l, s)=(48, 5)$.

\subsubsection{Dropout Parameter $\alpha$}
\begin{table}[ht]
    \centering
    \caption{Analysis on Interval-aware Dropout Parameter $\alpha$.}
    \label{tab:alpha}
    \begin{tabular}{c|cccc}\hline
    Type&$\alpha=0$&$\alpha=0.03$&$\bm{\alpha=0.1}$&$\alpha=0.2$\\\hline
    \multicolumn{5}{c}{\textbf{RMSE}}\\\hline
    Random&\textbf{0.1949}&0.2241&0.2082&0.2119\\
    Fiber&0.3466&0.3255&\textbf{0.3009}&0.3012\\
    Block&0.3579&0.3280&0.3090&\textbf{0.3046}\\\hline
    \multicolumn{5}{c}{\textbf{MAE}}\\\hline
    Random&\textbf{0.0969}&0.1024&0.1018&0.1001\\
    Fiber&0.1621&0.1533&0.1466&\textbf{0.1445}\\
    Block&0.1791&0.1600&\textbf{0.1482}&0.1517\\\hline
    \end{tabular}
\end{table}

Table \ref{tab:alpha} shows the sensitivity analysis on the hyperparameter $\alpha$ of the interval-aware dropout mechanism, which controls the decay rate of the dropout probability as defined in Equation (\ref{equ:dropout}): 
The analysis was performed by varying $\alpha$ across a range of values $(0, 0.03, 0.1, 0.2)$ while keeping the average dropout rate $p = 0.1$ fixed. Notably, when $\alpha=0$, the dropout probability equals $p$ for all the edges, and the mechanism degrades to standard dropout.

Two observations can be made based on Table \ref{tab:alpha}. First, in random missing scenarios, the imputation accuracy is not related to the parameter $\alpha$, or even the normal dropout with $\alpha=0$ sometimes obtains the best performance. This is because the imputation of random missing mainly relies on adjacent observed values, increasing the drop rate of them is useless for the overfitting problem. Second, in fiber and block missing conditions, the interval-aware dropout mechanism achieves a significant improvement (12.3\% in RMSE), proving its effectiveness in the assistance of learning long-term dependencies. Therefore, in PAST, the parameter $\alpha$ is set to 0.1 when the sequence length $L=96$, in order to alleviate the
overfitting problem while preventing the risk of underfitting of local dependencies.


\subsubsection{Multi-Order Convolution $K$}
\begin{table}[ht]
    \centering
    \caption{Analysis on Order $K$ in Spatial Layers.}
    \label{tab:k}
    \begin{tabular}{c|cccc}\hline
    Type & $K=0$ & $K=1$ & $\bm{K=2}$ & $K=3$ \\\hline
    \multicolumn{5}{c}{\textbf{RMSE}}\\\hline
    Random & 0.2974 & 0.2204 & 0.2082 & \textbf{0.2023} \\
    Fiber & 0.5690 & 0.3025 & \textbf{0.3009} & 0.3246 \\
    Block & 0.5502 & 0.3308 & \textbf{0.3090} & 0.3094 \\\hline
    \multicolumn{5}{c}{\textbf{MAE}}\\\hline
    Random & 0.1329 & 0.1169 & 0.1018 & \textbf{0.0987} \\
    Fiber & 0.2974 & 0.1489 & \textbf{0.1466} & 0.1599 \\
    Block & 0.2903 & 0.1562 & 0.1482 & \textbf{0.1459} \\\hline
    \end{tabular}
\end{table}

To investigate the influence of the multi-hop order $K$ in GIM, a sensitivity analysis is conducted by varying $K$ across the range of 0 to 3. Here, $K=0$ signifies that the model does not account for spatial associations. The experimental results are shown in Table \ref{tab:k}, which reveal that imputation accuracy progressively improves with increasing $K$, reflecting the critical role of enlarging the perception range. Accuracy gains are most pronounced from $K=0$ to $K=2$, which demonstrates effective capture of multi-hop convolutions. However, at $K=3$, improvements become marginal, with RMSE only slightly better than at $K = 2$ or even worse. Consequently, to optimize imputation effectiveness and efficiency, we recommend setting $K=2$ as the default value.

\subsubsection{Embedding Dimension $d$}
\begin{table}[ht]
    \centering
    \caption{Analysis on Embedding Dimension $d$.}
    \label{tab:d}
    \begin{tabular}{c|cccc}\hline
    Type&$d=16$&$d=32$&$\bm{d=64}$&$d=128$\\\hline
\multicolumn{5}{c}{\textbf{RMSE}}\\\hline
    Random&\textbf{0.2056}&0.2104&0.2082&0.2235\\
    Fiber&0.3384&0.3245&0.3009&\textbf{0.2967}\\
    Block&0.3464&0.3256&0.3090&\textbf{0.3028}\\\hline
    \multicolumn{5}{c}{\textbf{MAE}}\\\hline
    Random&\textbf{0.1009}&0.1017&0.1018&0.1222\\
    Fiber&0.1710&0.1592&\textbf{0.1466}&0.1493\\
    Block&0.1788&0.1619&0.1482&\textbf{0.1450}\\\hline
    \end{tabular}
\end{table}

The sensitivity analysis is conducted by varying $d$ across the range of 16 to 128. This dimension governs the representational capacity of embedded vectors in both GIM and CGM, directly affecting the model's ability to capture intricate spatio-temporal patterns.
The experimental results are shown in Table \ref{tab:d}, revealing that imputation accuracy improves progressively with increasing $d$, especially in fiber and block missing conditions. This is primarily because of the increased fitting capacity of PAST. However, the transition from $d = 64$ to $d = 128$ shows negligible improvement, suggesting that the model has sufficiently learned the dataset's underlying patterns.

Additionally, larger $d$ significantly degrades efficiency. To balance imputation effectiveness and efficiency, we recommend $d = 64$ for most scenarios. These findings highlight the importance of dimension selection to optimize performance without increasing computational burden.

\subsection{Case Study}
In this section, we visualize two specific imputation cases in the METR-LA dataset to exhibit the performance of PAST, and the contributions made by the graph-integrated module and the cross-gated module.

\begin{figure*}[t]
    \centering
    \includegraphics[width=\linewidth]{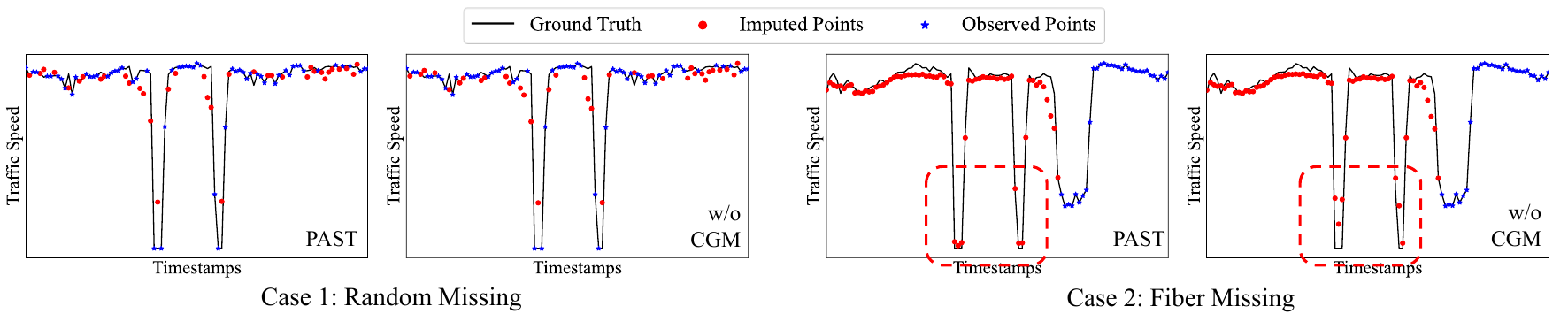}
    \caption{Two imputation cases on random and fiber missing scenarios.}
    \Description{}
    \label{fig:case}
\end{figure*}

As shown in Figure \ref{fig:case}, the blue points denote the observed values, while the red ones show the imputation results. The left figure in each case shows the imputation performance of PAST, while the right one shows the performance without the assistance of the cross-gated module (only imputed by the graph-integrated module). Based on case 1, we can observe that the imputation accuracy is high in both figures, which indicates that extracting primary patterns just based on GIM is sufficient for addressing random missing conditions. This is because the values at missing positions can be effectively inferred from the adjacent observations based on local dependencies.
In contrast, in case 2, where the first 60 points are consecutively missed (about 15 hours), the imputation results based on PAST is evidently better than those just based on GIM, especially when there are sudden drifts or pulses in the sequence. This implies that these missing values cannot be estimated by just learning relationships between data points, proving the necessity of employing the cross-gated module to extract auxiliary patterns in fiber missing conditions.

\section{Conclusion}
Traffic time series imputation is a foundational capability for intelligent transportation systems. The variety of missing types makes the imputation challenging. Effective solutions must reconcile multi-scale spatio-temporal dependencies with diverse missing patterns, and make principled use of internal and external information  to recover the incomplete structure.
This paper introduces PAST, a novel traffic time series imputation model that disentangles primary and auxiliary patterns, leveraging a graph-integrated module (GIM) and a cross-gated module (CGM) to complementarily impute missing values. Experimental results on MeTr-LA, PeMS-Bay, and LargeST-SD datasets across 27 missing conditions demonstrate PAST's superiority, outperforming seven baselines by up to 26.2\% in RMSE and 31.6\% in MAE, with enhanced stability in fiber and block missing scenarios. 

Future research will concentrate on two aspects: (1) The exploration of multimodal data integration. For example, we may look for fusion strategies of weather, events, holidays, or map semantics to enhance the auxiliary pattern extraction. (2) Real-time and adaptive imputation. Approaches such as streaming updates or concept drift detection will be employed to further improve performance in dynamic traffic contexts.

\balance

\bibliographystyle{ACM-Reference-Format}
\bibliography{Reference}

\end{document}